\documentclass{article}

\usepackage{microtype}
\usepackage{graphicx}
\usepackage{subcaption}
\usepackage{booktabs} % for professional tables

\usepackage{hyperref}

\usepackage[preprint]{icml2026}

\makeatletter
\renewcommand{\icmlcorrespondingauthor}[2]{%
  \gdef\icmlcorrespondingauthor@text{\@empty}%
}
\renewcommand{\printAffiliationsAndNotice}[1]{%
  \global\icml@noticeprintedtrue%
  \stepcounter{@affiliationcounter}%
  {\let\thefootnote\relax\footnotetext{\hspace*{-\footnotesep}\ificmlshowauthors #1\fi%
      \forloop{@affilnum}{1}{\value{@affilnum} < \value{@affiliationcounter}}{%
        \textsuperscript{\arabic{@affilnum}}\ifcsname @affilname\the@affilnum\endcsname%
          \csname @affilname\the@affilnum\endcsname%
        \else%
          {\bf AUTHORERR: Missing \textbackslash{}icmlaffiliation.}%
        \fi%
      }.%
      \ \\
      \Notice@String
    }%
  }%
}
\makeatother

\usepackage{amsmath}
\usepackage{amssymb}
\usepackage{mathtools}
\usepackage{amsthm}

\usepackage{amsfonts}
\usepackage{booktabs}              % booktabs线条
\usepackage{tabularx}              % 实现等宽列和占满页宽
\usepackage{multirow}              % 实现纵向单元格合并

\usepackage{xcolor}         % colors
\usepackage[most]{tcolorbox} % 放在导言区

\tcbuselibrary{breakable}
\usepackage[capitalize,noabbrev]{cleveref}

\theoremstyle{plain}
\newtheorem{theorem}{Theorem}[section]

\theoremstyle{definition}

\theoremstyle{remark}
\newtheorem{remark}[theorem]{Remark}
\usepackage{xcolor}

\usepackage{latexsym}
\usepackage{graphicx}
\usepackage{amsmath}
\usepackage{pifont}   % 提供勾和叉
\usepackage{xcolor}   % 提供颜色
\usepackage{amssymb} 
\usepackage{xcolor}
\usepackage{soul}
\usepackage{hyperref}
\usepackage{cleveref}
\usepackage{colortbl}
\usepackage{fontawesome}
\usepackage{pifont}

\definecolor{bestcell}{HTML}{F2F2FE}
\usepackage[textsize=tiny]{todonotes}

\icmltitlerunning{Explainable CoT Compression in Multimodal Large Reasoning Models}

\begin{document}

\twocolumn[
  \icmltitle{\textit{Bridging Efficiency and Transparency:} Explainable CoT Compression in Multimodal Large Reasoning Models}



  \icmlsetsymbol{equal}{*}

  \begin{icmlauthorlist}
    \icmlauthor{Yizhi Wang}{aaa,bbb}
    \icmlauthor{Linan Yue}{aaa,bbb}
    \icmlauthor{Min-Ling Zhang}{aaa,bbb}
  \end{icmlauthorlist}

  \icmlaffiliation{aaa}{School of Computer Science and Engineering, Southeast University}
  \icmlaffiliation{bbb}{Key Laboratory of Computer Network and Information Integration (SEU), Ministry of Education, China}

  \icmlcorrespondingauthor{}{}

  \icmlkeywords{Machine Learning, ICML}

  \vskip 0.3in
]



\printAffiliationsAndNotice{}  % no special notice (required even if empty)

\begin{abstract}
Long chains of thought (Long CoTs) are widely employed in multimodal reasoning models to tackle complex tasks by capturing detailed visual information. However, these Long CoTs are often excessively lengthy and contain redundant reasoning steps, which can hinder inference efficiency. Compressing these long CoTs is a natural solution, yet existing approaches face two major challenges: (1) they may compromise the integrity of visual–textual reasoning by removing essential alignment cues, and (2) the compression process lacks explainability, making it difficult to discern which information is critical. To address these problems, we propose XMCC, an eXplainable Multimodal CoT Compressor that formulates compression as a sequential decision-making process optimized via reinforcement learning. XMCC can effectively shorten reasoning trajectories while preserving key reasoning steps and answer correctness, and simultaneously generates natural-language explanations for its compression decisions. Extensive experiments on representative multimodal reasoning benchmarks demonstrate that XMCC not only reduces reasoning length but also provides explainable explanations, validating its effectiveness. Code is available at \url{https://github.com/Snowstorm1492/XMCC-Code}.
\end{abstract}

\begin{figure*}
    \centering
    \includegraphics[width=14cm]{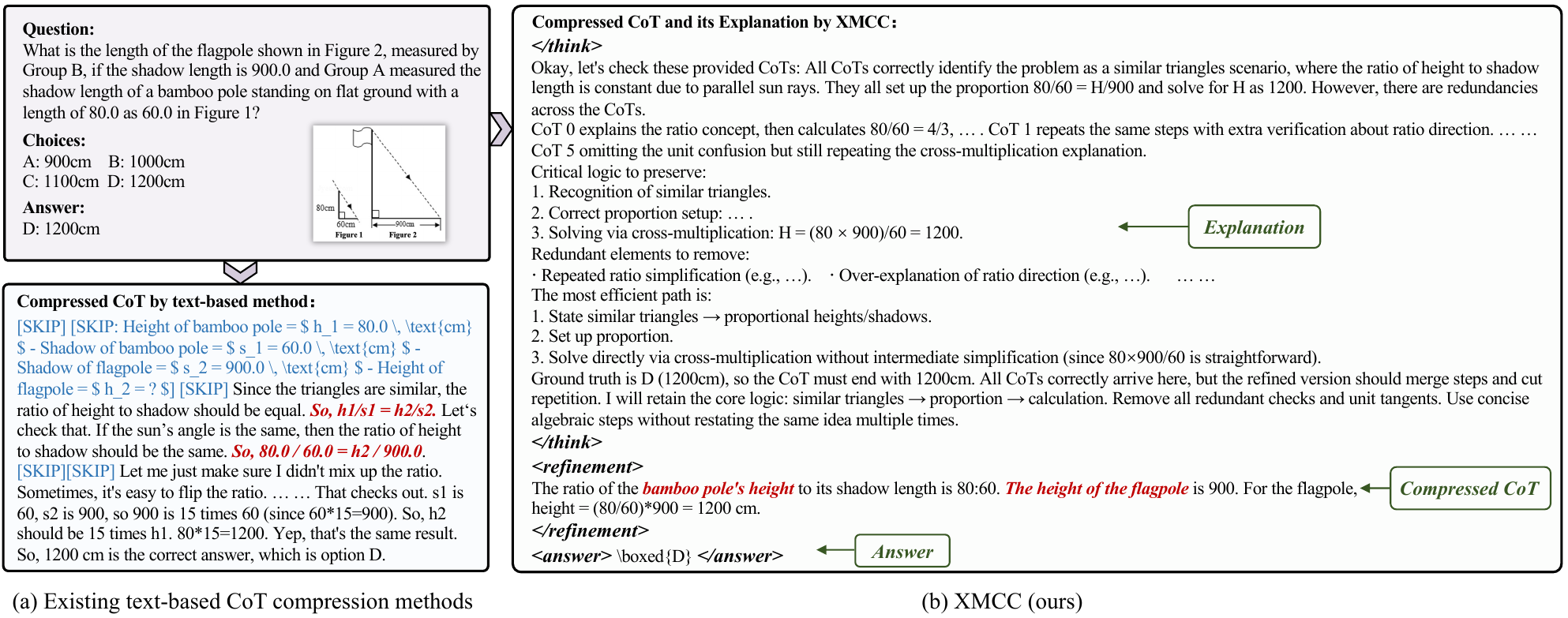}
    \caption{Differences between existing text-based CoT compression methods and XMCC. (a) shows the compressed CoT produced by a text-based compression method, while (b) shows the result from XMCC. In (a), each ``[SKIP]'' represents a deleted step. It can be observed that the text-based method erroneously removes critical visually grounded information that defines variable meanings (e.g., ``Height of bamboo pole = h\_1''). In contrast, XMCC preserves these critical alignment cues.}
    \label{fig:casestudy}
    \vspace{-0.5cm}
\end{figure*}

\section{Introduction}
Multimodal large reasoning models (MLRMs) have demonstrated remarkable advantages in solving complex reasoning tasks, typically by generating long chains of thoughts (Long CoTs) that contain rich descriptions of visual details, spatial relations, and vision–language alignments \cite{qwen3-vl, internvl3_5, kimi-vl, huang2025vision, yang2025r1, meng2025mm,zhang2024mm,zhou2025r1,zhang_mm15_2024}. However, in multimodal settings, such chains often involve repeated verbalization of low-level perceptual cues and cross-modal correspondences. This excessive accumulation of information leads to extremely long reasoning trajectories, which in turn severely limits reasoning efficiency in practical applications~\cite{zhang2026vskip,lee2025well,qu2025survey}.

To alleviate this issue, a straightforward approach is to directly adapt existing text-based CoT compression methods to the multimodal setting~\cite{cui2025stepwise,lu2025prolonged,chen2024not}. Specifically, one can first automatically compress the original multimodal CoTs, for example, by adopting segment-level compression strategies that identify and remove redundant or repetitive descriptive spans, thereby producing more concise reasoning trajectories. Based on these results, a multimodal CoT compression dataset can be constructed in the form of “\textit{original query–compressed CoT}” pairs, which is then used for supervised fine-tuning (SFT). This training enables the model to generate more compact multimodal CoTs at inference time, leading to substantial improvements in reasoning efficiency.

However, this naive transfer faces two fundamental challenges: (1) Conventional text-based compression methods tend to break the integrity of cross-modal reasoning~\cite{shen2025heima,zhang2026vskip}. In multimodal CoTs, certain textual descriptions not only convey semantics but also serve as spatial pointers and alignment cues for grounding visual evidence. 
For example, as illustrated in Figure \ref{fig:casestudy}(a), the original reasoning may contain statements such as “\textit{Height of bamboo pole = h\_1 = 80.0}”. If the compressor only preserves Algebraic expression like “\textit{h1/s1 = h2/s2}”, the removal of spatial localization makes it difficult for the model to correctly establish the vision–language correspondence.
Because it is hard to distinguish redundant verbal repetition from indispensable alignment cues, compression can easily disrupt the visual–textual reasoning chain.
(2) Existing approaches primarily optimize for high compression ratios while overlooking the explainability of the compression process. Models cannot explain why certain visual descriptions or intermediate steps are removed and which pieces of information are essential for the final conclusion. When key visual cues are discarded, users cannot tell whether this is due to true redundancy or to erroneous decisions made by the compressor, making the process a black box and limiting transparency and reliability.

Motivated by these issues, we aim to develop a compression mechanism that both preserves the logical integrity of multimodal reasoning and provides high transparency. Inspired by DeepSeek-R1, we formulate compression as a sequential decision-making process and optimize it with reinforcement learning (RL), specifically using the GRPO algorithm~\cite{shao2024deepseekmath}. By using downstream outcome accuracy as a posterior reward, the compressor is encouraged to automatically identify which visual descriptions and alignment information are most critical for correct reasoning, thereby significantly reducing chain length while retaining the most discriminative evidence for visual inference. Moreover, we introduce explicit reasoning traces in the RL framework.  Before producing the compressed CoT, the model is required to generate natural-language explanations for its compression decisions. As shown in Figure \ref{fig:casestudy}(b), these explanations are enclosed by \verb|<think>| tags, while the final compressed CoT is presented within \verb|<refinement>| tags.

Along this research line, in this paper, we propose an e\textbf{X}plainable \textbf{M}ultimodal \textbf{C}oT \textbf{C}ompressor (\textbf{XMCC}) that improves reasoning efficiency while performing transparent compression of multimodal CoTs. Specifically, we first leverage MLRMs to automatically synthesize Long CoTs, thereby constructing a multimodal CoT dataset to be compressed. We then design a multi-component RL reward function tailored for CoT compression, which guides the model to eliminate redundant reasoning steps while preserving answer correctness. The reward function consists of four key components: \textit{a format reward}, \textit{an outcome reward}, \textit{a step-wise criticality reward}, and \textit{a length penalty reward}. Among them, the format reward enforces a structured generation order of “\textit{explanation–compressed CoT–final answer}”, ensuring controllability and readability of the compression process. The outcome reward requires that the compressed CoTs still support the model in producing the correct answer, thereby maintaining semantic and logical validity. Based on the two common rewards, the step-wise criticality reward performs a fine-grained evaluation of each segment in the compressed CoTs by measuring its contribution to task performance. This allows the model to distinguish redundant steps from indispensable ones, retaining only those reasoning components that are truly essential for the final outcome. Meanwhile, the length penalty adapts to the complexity of each query by characterizing it through the length of the original CoTs, and accordingly adjusts the compression strength to avoid over-compression for simple cases or under-compression for complex ones. Under the joint optimization of these rewards, the compressor is able to substantially shorten reasoning trajectories while effectively preserving the core logical structure and alignment cues required for multimodal reasoning. Finally, the compressed multimodal CoTs are used to SFT the base model, resulting in an efficient multimodal reasoner that achieves both high inference efficiency and strong task performance.

Our main contributions are summarized as follows:

$\bullet$ We propose XMCC, an explainable multimodal CoT compression framework that significantly shortens reasoning trajectories while preserving key visual evidence, and provides natural-language explanations for its compression decisions

$\bullet$ We formulate multimodal CoT compression as a sequential decision-making problem and design a multi-component RL reward function, enabling fine-grained identification of indispensable reasoning steps.

$\bullet$ Extensive experiments on representative multimodal reasoning benchmarks demonstrate that XMCC consistently achieves substantial reductions in CoT length while maintaining or even improving task accuracy, validating the effectiveness of our proposed XMCC.
% \begin{itemize}\setlength{\itemsep}{0pt}\setlength{\parskip}{0pt}
%   \item We propose XMCC, an explainable multimodal CoT compression framework that significantly shortens reasoning trajectories while preserving key visual evidence, and provides natural-language explanations for its compression decisions.
%   \item We formulate multimodal CoT compression as a sequential decision-making problem and design a multi-component RL reward function, enabling  fine-grained identification of indispensable reasoning steps.
%   \item Extensive experiments on representative multimodal reasoning benchmarks demonstrate that XMCC consistently achieves substantial reductions in CoT length while maintaining or even improving task accuracy, validating the effectiveness of our proposed XMCC.
% \end{itemize}

\section{Related Work}

\subsection{Multimodal Reasoning}
Recently, Multimodal Large Reasoning Models (MLRMs) have demonstrated remarkable capabilities in tackling complex visual reasoning tasks~\cite{qwen3-vl, wang2025skywork, xie_logic-rl_2025,shen2025vlm,chen2025advancing,step3blog,wu_deepseek-vl2_2024,wang_enhancing_2025,zhang_improve_2024}. Early endeavors, such as LLaVA-CoT~\cite{xu_llava-cot_2025} and Mulberry~\cite{yao_mulberry_2024}, explored the feasibility of structured multimodal reasoning through prompt engineering or Monte Carlo Tree Search. Following the advent of DeepSeek-R1~\cite{guo2025deepseek}, researchers began incorporating RL-based post-training into multimodal models, enhancing multimodal reasoning capabilities. Lately, a series of advanced MLRMs have further pushed the boundaries of this field~\cite{qwen3-vl, internvl3_5, kimi-vl, li_llava-onevision_2024,qvq-72b-preview}. Their generated CoTs can finely integrate visual perception with logical deduction, achieving good performance on challenging tasks such as multi-image understanding.

However, such performance gains often come at the expense of efficiency. Compared to textual reasoning, the verbosity of multimodal CoTs is particularly pronounced, partly due to their inherent cross-modal interaction mechanisms. In practice, MLRM-generated CoTs often contain repetitive visual descriptions and unnecessary self-reflective statements. Such content frequently renders multimodal CoTs excessively long, thereby degrading inference efficiency.

\subsection{CoT Compression}
To alleviate inference overhead, CoT compression has gained increasing attention. Its core objective is shortening CoTs while preserving logic critical to the final answer~\cite{hu2026conmax,xu2025thought,zhuang2025accelerating}. 
Early methods primarily relied on prompt engineering, using directive constraints (e.g., ``use at most $k$ tokens''~\cite{han2024token}) to encourage more concise CoTs~\cite{renze2024benefits,nayab2024concise}. While simple to implement, they suffer from limited generalization and struggle to adapt to diverse task complexities.

Recent research has shifted toward data-driven compression paradigms, which can be categorized by operation granularity into token-level and step/block-level compression~\cite{cui2025stepwise,yuan2025not}. At the token level, researchers prune tokens based on estimated information contribution. For example, \citet{yuan2025not} proposed Conditional Token Selection (CTS), which dynamically estimates token importance using the perplexity of a reference model. \citet{xia2025tokenskip} introduced TokenSkip, retaining only tokens whose importance exceeds a predefined threshold. At a higher semantic level, step-wise compression segments the reasoning chain into coherent units and selectively preserves key steps. For instance, \citet{xiao2025limopro} proposed a Perplexity-based Importance Refinement (PIR) framework to differentiate the contributions of individual reasoning steps. \citet{wang2025r1} generate multiple simplified candidates for each semantic block and apply a greedy strategy to balance conciseness and fidelity. Additionally, some works attempt structural reorganization: \citet{zhao2025can} converted CoTs into logical graphs and pruned low-value nodes to achieve structural compression.

While these methods have proven effective in pure-text scenarios, extending them to the multimodal setting presents unique challenges. First, existing approaches predominantly rely on internal linguistic signals (e.g., perplexity) for compression decisions and fail to adequately account for the integrity of cross-modal grounding.
Second, current compression processes lack interpretability. Users cannot discern the rationale behind pruning decisions. This lack of transparency limits their applicability in high-assurance or safety-critical settings.

\section{Explainable Multimodal CoT Compressor}

\subsection{Overview of the Compressor}
\begin{figure*}[t]
    \centering
    \includegraphics[width=16.cm]{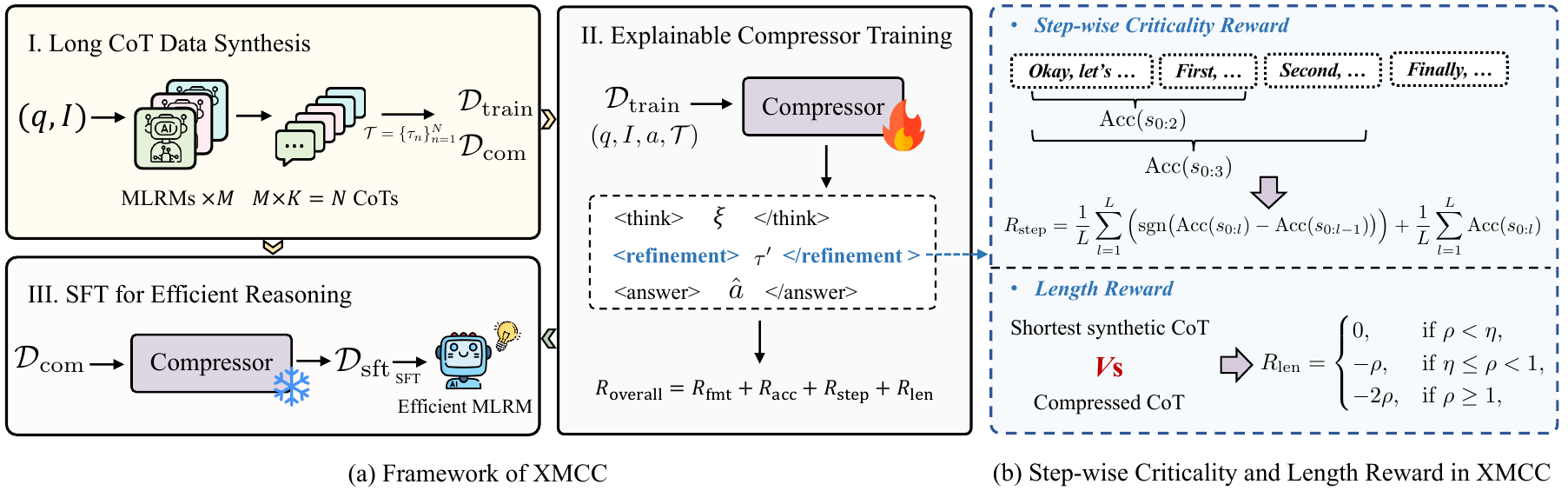}
    \caption{Overview of XMCC. (a) The framework consists of three stages: (I) synthesizing diverse long CoTs from heterogeneous MLRMs; (II) training an explainable compressor via RL with the proposed reward function; and (III) SFT on compressed CoTs for efficient inference. (b) In the proposed reward function, step-wise criticality reward evaluating each segment's contribution to task performance, to ensure the quality of compressed reasoning. The length reward adapts compression intensity to task complexity.}
    \label{fig:model}
    \vspace{-0.2cm}
\end{figure*}

\textbf{Problem Formulation.}
We study the problem of CoT compression in multimodal reasoning. Given a textual query $q$, a corresponding image $I$, the answer $a$, and a set of original long CoT trajectories $\mathcal{T} = \{\tau_n\}_{n=1}^N$ (which may contain a single trajectory or multiple diverse CoTs generated for the same $(q, I)$ pair), our goal is to train a compressor $f_{\theta}(\cdot)$ that can learn an explainable compression mapping:
\begin{equation}
\mathcal{F}: (q, I, a, \mathcal{T}) \longmapsto (\xi,  \tau', \hat{a}),
\end{equation}
where $\xi$ is a natural-language explanation of the compression rationale, justifying pruning decisions. $\tau'$ denotes the compressed CoT, which significantly shortens the reasoning trajectory while preserving sufficient reasoning capability. $\hat{a}$ is the predicted answer. Although the ground-truth answer $a$ is provided as part of the input, we explicitly require the model to output a prediction $\hat{a}$. This design ensures that the compressor performs compression under the guidance of the ground-truth answer, thereby enhancing the correctness and quality of the refined CoT.

Through this compression, our aim is to generate CoTs that are both concise and logically complete, while providing natural-language explanations for the CoT compression.

\textbf{Overall Framework.}
As illustrated in Figure~\ref{fig:model}(a), we propose a RL-driven eXplainable Multimodal CoT Compressor (XMCC). The compressor is designed to significantly shorten inference trajectories while preserving critical visual-language alignment cues, and to provide explainable compression decisions.
Specifically, the training of XMCC proceeds in three stages: 

\textbf{({I}) Long CoT Data Synthesis:} We first leverage MLRMs to generate long CoTs, forming the dataset $\mathcal{D}_{\text{train}}$ for compression training and the dataset to be compressed $\mathcal{D}_{\text{com}}$. 

\textbf{({II}) Explainable Compressor Training:} We then employ a lightweight multimodal reasoning model as the CoT compressor $f_\theta$, modeling the compression process as a sequential decision-making problem. The compressor takes a quadruple $(q, I, a, \mathcal{T})$ as input and is trained via RL using GRPO~\cite{shao2024deepseekmath}. During training, the compressor is guided by a composite reward signal that balances compression efficiency and robustness. The reward consists of four components: 
\textit{format},
\textit{outcome}, \textit{step-wise criticality}, and \textit{length penalty reward}. Among them, format and outcome rewards ensure the correctness of the final answer and structural compliance, while step-wise criticality and length rewards guide the policy from the perspectives of reasoning quality and trajectory efficiency.

Through this RL framework, the compressor learns not only \textit{how to compress}  but also \textit{why to compress}, retaining the most discriminative visual-language reasoning signals while simultaneously generating natural-language explanations for each compression decision. 

\textbf{({III}) SFT for Efficient Reasoning:} By applying the compressor to the original dataset $\mathcal{D}_{\text{com}}$, we obtain a compressed dataset $\mathcal{D}_{\text{sft}}$ that consists of the original $(q, I)$ pairs and the compressed trajectories $\tau'$. Supervised fine-tuning on $\mathcal{D}_{\text{sft}}$ enables the model to perform more efficient reasoning.

\subsection{Long CoT Data Synthesis}
\label{sec:syntrajs}
In this section, we present the inputs to the CoT compressor. Specifically, the inputs consist of a query $q$, an image $I$, the ground-truth answer $a$, and long CoTs. Among them, $q$, $I$, and $a$ can be directly obtained from the dataset, while the CoTs are generated by the multimodal reasoning models.
To achieve CoTs, a straightforward approach is to use a multimodal model to generate a single CoT for each sample. However, relying on a single trajectory is inherently limited, since even when the final prediction is correct, the intermediate reasoning may still contain local errors. When such a trajectory is fed into the compressor, these defects can be propagated to the compressed CoT, thereby degrading its logical quality and information density. To improve the robustness and reliability of the input, we adopt a ``\textit{multi-model, multi-sampling}'' strategy. Specifically, for each $(q, I)$, we employ $M$ heterogeneous multimodal reasoning models and sample $K$ CoTs from each model, yielding in total $N = M \times K$ distinct long CoTs that form the set $\mathcal{T} = \{\tau_n\}_{n=1}^N$.
This set is provided as the input to the CoT compressor, which performs inductive and contrastive compression over multiple CoTs to produce a more compact and reliable reasoning chain.

Finally, the synthesized data are split into two parts: one forms the dataset $\mathcal{D}_{\text{train}}$ for training the compressor, and the other serves as the dataset $\mathcal{D}_{\text{com}}$ to be compressed, which is used for subsequent SFT.

\remark
The key motivation behind this design is to leverage the natural diversity of CoTs produced by different models. Reasoning steps that consistently appear across multiple CoTs are more likely to be essential for solving the task, whereas model-specific or conflicting steps often stem from biases or stochastic noise. Therefore, retaining high-consensus components while discarding divergent ones enables the generation of compressed CoTs that are both concise and logically reliable.

\subsection{Explainable Compressor Training}
\subsubsection{Reward Design}

To guide the compressor in shortening CoT length while preserving visual-language alignment cues and high-order logical structures essential for multimodal reasoning, we design a composite reward function composed of four signals. Beyond ensuring task validity and output compliance, this reward system introduces two novel components (i.e., step-wise criticality reward and length reward) to enable fine-grained control and adaptive regulation of compression quality. The former evaluates the contribution of each reasoning unit from a content perspective, while the latter dynamically balances compression intensity against task complexity from an efficiency perspective, jointly providing supervision over the compression behavior.

\textbf{Format Reward.}
To ensure that the compressor produces structured and machine-parseable explanations, we impose a format constraint via the format reward $R_{\text{fmt}}$. A valid output is required to follow a fixed template: it begins with a reasoning block enclosed by the \texttt{<think>} tag, which provides a natural-language explanation $\xi$ of the compression process. The resulting compressed CoT $\tau'$ is placed within the \texttt{<refinement>} and \texttt{</refinement>} tags. Finally, the predicted answer $\hat{a}$ is wrapped by the \texttt{<answer>} tag. The format reward assigns a binary signal by checking the presence and correctness of these tags:
\begin{equation}
R_{\text{fmt}} = 
\begin{cases}
  1, & \text{if the output format is valid}, \\
  0, & \text{otherwise}.
\end{cases}
\end{equation}
This design enforces a unified output structure and facilitates reliable parsing and evaluation of both the compressed reasoning and the final prediction.

\textbf{Outcome Reward.}
As a foundational constraint, the accuracy reward $R_{\text{acc}}$ verifies  whether the predicted answer $\hat{a}$ exactly matches the ground-truth label $a$, using regular-expression-based parsing:
\begin{equation}
R_{\text{acc}} = \mathbb{I} \left[ \hat{a} = a \right],
\end{equation}
where $\mathbb{I}[\cdot]$ denotes the indicator function. The purpose of this reward design is to encourage the compressor to preserve the correctness of the reasoning, ensuring that the compressed CoT can lead to the correct final answer.

\textbf{Step-wise Criticality Reward.}
As illustrated in Figure \ref{fig:model}(b), in this section, we introduce a step-wise criticality reward $R_{\text{step}}$ for fine-grained quantification of compressed CoTs, aiming to measure the effectiveness of each step in the chain of thought and thereby guide the compression process.
 Specifically, we partition the compressed trajectory $\tau'$ into $L$ semantically coherent segments $\{s_l\}_{l=1}^L$. For notational convenience, we define $s_0 = \varnothing$ (the empty sequence) and denote by $s_{0:l}$ the concatenation of segments from $s_0$ through $s_l$, (i.e., $s_{0:l} = (s_1, s_2, \cdots, s_l)$). For each segment $s_l$, we construct an input $(q, I, s_{0:l})$ and feed it into a lightweight multimodal verification model (e.g., Qwen3-VL-2B-Instruct) which does not equip complex reasoning capabilities. Thus, if it can correctly answer the question with the aid of $s_{0:l}$, we interpret $s_{0:l}$ as containing effective reasoning content. To assess this, we perform independent inference runs with three different random seeds and compute the average matching accuracy between the predicted answers and the ground-truth $a$, denoted as $\text{Acc}(s_{0:l})$. The step-wise criticality reward is then defined as:
\begin{equation}
\small
R_{\text{step}} = \frac{1}{L} \sum_{l=1}^L \Big( \text{sgn}\big( \text{Acc}(s_{0:l}) - \text{Acc}(s_{0:l-1}) \big) \Big) + \frac{1}{L} \sum_{l=1}^L \text{Acc}(s_{0:l}),
\end{equation}
which consists of two components: an \textit{accuracy gain reward} and an \textit{accuracy reward}. The first term, the \textit{accuracy gain reward}, employs the sign function $\text{sgn}(\cdot)$ which takes the value 1 when the input is greater than 0, and 0 otherwise.
This term measures whether the inclusion of segment $s_l$ contributes positively to task performance. If $\text{Acc}(s_{0:l}) > \text{Acc}(s_{0:l-1})$, the segment $s_l$ is deemed beneficial and receives a reward of 1; otherwise, it receives no reward, indicating that $s_l$ is either redundant or potentially harmful (e.g., introducing noise or errors).

While the \textit{accuracy gain reward} effectively captures marginal contributions of individual segments, it may be insufficient when the sequence $\{\text{Acc}(s_{0:l})\}_{l=1}^L$ exhibits high variance. This could lead to situations where a low-quality CoT receives an inflated reward due to sporadic gains. To address this, we introduce the second term, the \textit{accuracy reward}, which computes the average accuracy across all sub-sequences $\{s_{0:l}\}_{l=1}^L$. This term provides a holistic assessment of the overall reasoning quality and mitigates potential reward hacking.

\remark
These two terms complement each other: the \textit{accuracy gain reward} emphasizes incremental utility, while the \textit{accuracy reward} ensures overall quality. Their combination yields a comprehensive and robust supervisory signal for evaluating the quality of compressed CoTs.

\textbf{Length Reward.}
Beyond ensuring logical completeness and step criticality, we impose reasonable constraints on output length to achieve truly efficient inference. However, the intrinsic complexity of visual reasoning tasks varies widely: simple tasks (e.g., object counting) may require only one or two sentences, whereas complex ones (e.g., multi-step spatial reasoning or cross-image comparison) inherently demand longer chains. A uniform length penalty would either under-compress simple tasks (wasting computation) or over-compress complex ones (harming reasoning fidelity), destabilizing overall performance.

To address this, we propose a \textit{difficulty-aware dynamic length regulation} mechanism, using the length of the original CoT as a proxy for task complexity to adaptively modulate compression intensity. Specifically, let $\min_{\tau \in \mathcal{T}}(\text{len}(\tau))$ denote the length of the shortest input CoT, and $\text{len}(\tau')$ the length of the compressed CoT. We define the compression ratio as $\rho = \tfrac{\text{len}(\tau')}{\min_{\tau \in \mathcal{T}}(\text{len}(\tau))}$. The dynamic length penalty is then formulated as:
\begin{equation}
R_{\text{len}} = 
\begin{cases}
  0, & \text{if } \rho < \eta, \\
  -\rho,   & \text{if } \eta \leq \rho < 1, \\
  -2\rho, & \text{if } \rho \geq 1,
\end{cases}
\end{equation}
where $0 < \eta < 1$ is a hyperparameter controlling the penalty threshold. This piecewise function implements a clear regulation logic: when the CoT has already been compressed into the target range ($\rho < \eta$), no penalty is applied to preserve reasoning integrity for complex tasks. When the compressed CoT is shorter than the shortest input trajectory ($\eta \leq \rho < 1$), a linear penalty is applied to encourage conciseness. Finally, when the length of the compressed CoT is greater than or equal to that of the shortest input ($\rho \geq 1$), the output is deemed unnecessarily verbose, so a doubled penalty is imposed to suppress such unreasonable behavior. Through this mechanism, the model learns to ``tailor its compression strategy to task difficulty,'' achieving an optimal trade-off between inference efficiency and accuracy.

In summary, our total reward function integrates four objectives. It can be formally expressed as:
\begin{equation}
\label{overall-eq}
R_{\text{overall}} = R_{\text{fmt}} +  R_{\text{acc}} + R_{\text{step}} + R_{\text{len}}.
\end{equation}
This design enables the compressor to actively eliminate redundancy while preserving critical reasoning logic and simultaneously generating explainable justifications, thereby laying a foundation for efficient, reliable, and trustworthy multimodal reasoning.

\subsubsection{RL Training with GRPO}
Based on the above reward design, we employ Group Relative Policy Optimization (GRPO)~\cite{shao2024deepseekmath}as our RL algorithm to post-train the compressor (a lightweight MLRM). Guided by the multi-component reward signal, it gradually learns a compression policy that bridges efficiency and transparency. The GRPO objective is formulated as:
\begin{equation}
\begin{aligned}
J_{\text{GRPO}}(\theta) &= \mathbb{E} \Bigg[ \frac{1}{G} \sum_{i=1}^{G} \Bigg( \min\Bigg( \frac{\pi_\theta(o_i|q)}{\pi_{\theta_{\text{old}}}(o_i|q)} A_i, \\
&\quad \operatorname{clip}\left( \frac{\pi_\theta(o_i|q)}{\pi_{\theta_{\text{old}}}(o_i|q)}, 1-\varepsilon, 1+\varepsilon \right) A_i \Bigg) \\
&- \beta \, \mathbb{D}_{\mathrm{KL}}\big( \pi_\theta(\cdot|q) \,\|\, \pi_{\text{ref}}(\cdot|q) \big) \Bigg) \Bigg],
\end{aligned}
\end{equation}
where the advantage function $A_i$ is computed based on an overall reward function $R_{\text{overall}}(\cdot)$ that reflects the holistic quality of the compressed output.

\subsection{SFT for Efficient Reasoning}
After training the compressor, we apply it to the original dataset $\mathcal{D}_{\text{com}}$ to obtain the refined dataset $\mathcal{D}_{\text{sft}}$, where each sample consists of the original query–image–answer triple $(q, I, a)$ and the corresponding compressed CoT $\tau'$. We then perform supervised fine-tuning (SFT) on a base model using $\mathcal{D}_{\text{sft}}$ with the standard next-token prediction loss. Through SFT on this refined dataset, we obtain a model capable of highly efficient reasoning. During inference, the model directly generates concise CoTs, significantly improving efficiency while maintaining solid task performance.

\section{Experiments}

\begin{table*}[t!]
\centering
\caption{Performance comparison on four multimodal reasoning benchmarks. Accuracy (Acc. $\uparrow$) measures effectiveness, while the average CoT length (AvgLen $\downarrow$) reflects reasoning efficiency. Ratio ($\downarrow$) denotes AvgLen divided by Acc.}
   \renewcommand\arraystretch{.95}
\setlength{\tabcolsep}{3.3mm}{
    \scalebox{0.8}{
\begin{tabular}{lcccccccccccc}
\toprule
\multirow{2}{*}{Method} & 
\multicolumn{3}{c}{\textbf{MathVista}} & 
\multicolumn{3}{c}{\textbf{WeMath}} & 
\multicolumn{3}{c}{\textbf{MMStar}} & 
\multicolumn{3}{c}{\textbf{MMMU}} \\
\cmidrule(lr){2-4} \cmidrule(lr){5-7} \cmidrule(lr){8-10} \cmidrule(lr){11-13}
 & Acc. & AvgLen & Ratio & Acc. & AvgLen & Ratio & Acc. & AvgLen & Ratio & Acc. & AvgLen & Ratio \\
\midrule
\multicolumn{6}{l}{\small{\textit{\textbf{SFT on Qwen2.5-VL}}}} \\
No Compression  & 35.8 & 524.1 & 14.6 & \textbf{55.8} & 572.5 & 10.3 & 55.1 & 452.7 & 8.2 & \textbf{50.4} & 565.7 & 11.2 \\
Prune-on-Logic  & 37.3 & 324.2 &  8.7 & 55.4 & 460.5 &  8.3 & 56.2 & 248.0 & 4.4 & 48.4 & 414.9 & 8.6 \\
StepEntropy     & 36.8 & 232.2 &  6.3 & 54.6 & 334.2 &  6.1 & 54.1 & 211.1 & 3.9 & 48.5 & 281.3 & 5.8 \\
\rowcolor{bestcell!90} XMCC            & \textbf{37.6} &  \textbf{89.7} &  \textbf{2.4} & 54.9 &  \textbf{99.6} &  \textbf{1.8} & \textbf{56.4} &  \textbf{87.5} & \textbf{1.6} & 49.2 & \textbf{105.8} & \textbf{2.2} \\
\midrule
\multicolumn{6}{l}{\small{\textit{\textbf{SFT on Qwen3-VL}}}} \\
No Compression  & 45.7 & 511.6 & 11.2 & \textbf{63.8} & 562.6 &  8.8 & 58.3 & 422.6 & 7.2 & 55.2 & 558.9 & 10.1 \\
Prune-on-Logic  & \textbf{45.9} & 299.5 &  6.5 & 63.4 & 419.5 &  6.6 & \textbf{57.7} & 229.0 & 4.0 & 55.7 & 382.4 & 6.9 \\
StepEntropy     & 43.9 & 251.3 &  5.7 & 61.6 & 313.5 &  5.1 & 58.1 & 225.5 & 3.9 & 54.4 & 270.5 & 5.0 \\
\rowcolor{bestcell!90}  XMCC            & 45.3 &  \textbf{91.0} &  \textbf{2.0} & 63.0 &  \textbf{97.7} &  \textbf{1.6} & \textbf{59.0} &  \textbf{86.8} & \textbf{1.5} & 54.8 & \textbf{107.3} & \textbf{2.0 }\\
\bottomrule
\end{tabular}
    }
    }  
\label{tab:main_results1}
\end{table*}

\begin{table*}[t!]
\centering
\caption{Performance comparison on R1-Onevision-Bench. Accuracy (Acc. $\uparrow$) measures effectiveness, while the average CoT length (AvgLen $\downarrow$) reflects reasoning efficiency. Ratio ($\downarrow$) denotes AvgLen divided by Acc.}
   \renewcommand\arraystretch{.95}
\setlength{\tabcolsep}{3.3mm}{
    \scalebox{0.8}{
\begin{tabular}{lcccccccccccc}
\toprule
\multirow{2}{*}{Method} & 
\multicolumn{3}{c}{\textbf{Deduction}} & 
\multicolumn{3}{c}{\textbf{Math}} & 
\multicolumn{3}{c}{\textbf{Physics}} & 
\multicolumn{3}{c}{\textbf{Overall}} \\
\cmidrule(lr){2-4} \cmidrule(lr){5-7} \cmidrule(lr){8-10} \cmidrule(lr){11-13}
 & Acc. & AvgLen & Ratio & Acc. & AvgLen & Ratio & Acc. & AvgLen & Ratio & Acc. & AvgLen & Ratio \\
\midrule
\multicolumn{6}{l}{\small{\textit{\textbf{SFT on Qwen2.5-VL}}}} \\
No Compression  & \textbf{27.8} & 834.3 & 30.0 & \textbf{25.7} & 629.5 & 24.5 & 23.4 & 748.0 & 32.0 & \textbf{29.1} & 656.2 & 22.5 \\ 
Prune-on-Logic  & 27.3 & 588.3 & 21.5 & 25.4 & 418.7 & 16.5 & 25.2 & 513.2 & 20.4 & 28.7 & 446.9 & 15.6 \\ 
StepEntropy     & 26.5 & 522.2 & 19.7 & 24.9 & 399.0 & 16.0 & 26.6 & 396.4 & 14.9 & 28.0 & 380.0 & 13.6 \\ 
\rowcolor{bestcell!90}  XMCC            & 27.6 & \textbf{115.0} & \textbf{4.2}  & 25.4 & \textbf{121.2} & \textbf{4.8}  & \textbf{28.1} & \textbf{116.8} & \textbf{4.2}  & 28.9 & \textbf{111.3} & \textbf{3.9}  \\ 
\midrule
\multicolumn{6}{l}{\small{\textit{\textbf{SFT on Qwen3-VL}}}} \\
No Compression  & 21.4 & 819.0 & 38.3 & 24.8 & 617.7 & 24.9 & 33.1 & 697.1 & 21.1 & 33.8 & 621.2 & 18.4 \\ 
Prune-on-Logic  & \textbf{23.0} & 602.6 & 26.2 & \textbf{25.0} & 428.9 & 17.2 & 34.4 & 510.3 & 14.8 & \textbf{34.1} & 442.6 & 13.0 \\ 
StepEntropy     & 22.5 & 560.0 & 24.9 & 22.9 & 397.7 & 17.4 & 31.7 & 432.3 & 13.6 & 32.0 & 384.9 & 12.0 \\ 
\rowcolor{bestcell!90}  XMCC            & 22.8 & \textbf{103.7} & \textbf{4.5}  & 24.5 & \textbf{120.6} & \textbf{4.9}  & \textbf{34.9} & \textbf{114.6} & \textbf{3.3}  & 33.3 & \textbf{112.9} & \textbf{3.4}  \\
\bottomrule
\end{tabular}
}}
\label{tab:main_results2}
\vspace{-0.4cm}
\end{table*}

\subsection{Experiment Settings}

\textbf{Datasets.}
To comprehensively evaluate the impact of CoT compression on multimodal reasoning, we introduce XMCC-Dataset, which contains 9,000 samples. This dataset aggregates instances from Geo170k~\cite{gao2023g} and ScienceQA~\cite{saikh2022scienceqa}, spanning diverse domains including mathematical reasoning, geometric analysis, and scientific reasoning. Each sample consists of an image-query-answer triplet paired with $N=6$ distinct long chains of thought (CoTs). These CoTs are generated by $K=3$ heterogeneous MLRMs (namely, Qwen3-VL~\cite{qwen3-vl}, InternVL3.5~\cite{internvl3_5}, and Kimi-VL~\cite{kimi-vl}) using the trajectory synthesis method described in Section~\ref{sec:syntrajs}.

\textbf{Comparison Methods.}
As explainable reasoning compression remains an emerging and underexplored research direction, the body of directly comparable prior work remains quite limited, particularly in multimodal settings. To establish meaningful baselines, we adapt two state-of-the-art (SoTA) CoT compression methods from the textual domain: Prune-on-Logic~\cite{zhao2025can} and StepEntropy~\cite{li2025compressing}. Specifically, {Prune-on-Logic} converts a CoT into a reasoning graph via semantic parsing and prunes nodes based on perplexity changes upon removal. {StepEntropy} segments the CoT into steps, computes step-level entropy from token-wise predictive uncertainty, and retains high-entropy (i.e., less redundant) steps for compression.

\begin{figure*}[t]
    \centering
    \includegraphics[width=12.5cm]{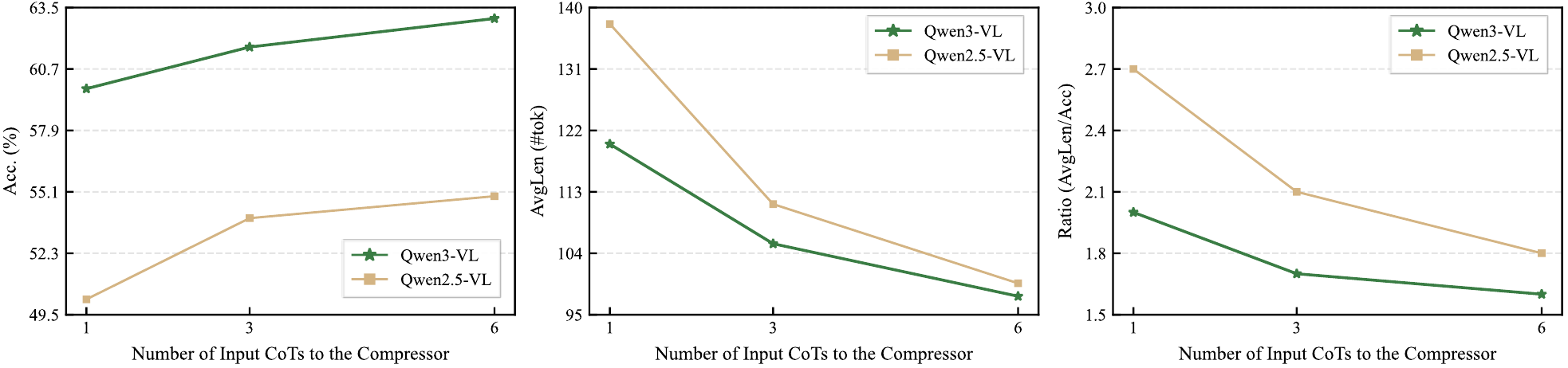}
    \caption{Analysis of input CoT quantity. From left to right: model accuracy, average reasoning length, and the accuracy-to-length ratio as functions of the number of input CoTs. As shown, increasing the number of input CoTs improves both task performance and efficiency.}
    \label{fig:combo}
    \vspace{-0.3cm}
\end{figure*}

\textbf{Implementation Details.}
We adopt Qwen3-VL-2B as the base architecture for our compressor, considering its lightweight parameter count and strong multimodal understanding capabilities. Through comparative experiments, we ultimately set the hyperparameter $\eta=0.15$. For supervised fine-tuning (SFT), we utilize Qwen2.5-VL-7B and Qwen3-VL-8B as the backbones, respectively. All SFT procedures are implemented using the LLaMA-Factory framework~\cite{zheng2024llamafactory}, which offers a unified and flexible interface for instruction tuning.

During evaluation, we assess model performance across several widely recognized multimodal reasoning benchmarks: MathVista~\cite{lu2023mathvista}, WeMath~\cite{qiao2024we}, MMStar~\cite{chen2024mmstar}, MMMU~\cite{yue2024mmmu}, and R1-Onevision-Bench~\cite{yang2025r1}. These datasets encompass a broad range of visual reasoning tasks with varying levels of complexity and modality interactions. We report two evaluation metrics: \textbf{Accuracy} and \textbf{Average CoT Length}. Accuracy reflects the effectiveness of the reasoning, while average CoT length measures its efficiency.

\subsection{Main Results}

To evaluate the effectiveness of XMCC, we conduct experiments on a suite of multimodal reasoning benchmarks. The results are summarized in Table~\ref{tab:main_results1} and~\ref{tab:main_results2}. We report the ``Ratio'' metric, defined as the average reasoning length divided by accuracy, to jointly assess both efficiency and performance. As shown, models trained on data refined by XMCC consistently outperform existing SoTA methods in terms of efficiency. In terms of task accuracy, XMCC matches or even surpasses SoTA approaches while using significantly fewer tokens. For example, on the Physics subset of R1-Onevision-Bench, the model trained with XMCC-refined data achieves an accuracy improvement of $2\%\!-\!3\%$ over the best baseline, while using only about $1/3$ of the tokens required by the SoTA. This demonstrates that XMCC can achieve substantial compression without compromising reasoning performance, and in some cases even improves it.

\begin{table}[t]
\centering
\caption{Evaluation of visual information preservation capability and natural-language explanation quality, where No Training refers to the untrained compressor of XMCC (i.e., the base model without CoT compression training).
``$\--$'' indicates that the method offers no explainability.}
   \renewcommand\arraystretch{1.2}
\setlength{\tabcolsep}{2.8mm}{
    \scalebox{0.8}{
\begin{tabular}{lcc}
\toprule
Method & Visual Infor. $\uparrow$ & Explainability $\uparrow$\\
\midrule
Prune-on-Logic & 3.51 & \--- \\
Step Entropy   & 2.83 &  \--- \\
No Training    & 3.47 & 2.45 \\
\rowcolor{bestcell!90} XMCC           & \textbf{3.95} & \textbf{4.17} \\
\bottomrule
\end{tabular}
   }
    }   
\label{tab:vis-expl}
\vspace{-0.4cm}
\end{table}

\subsection{Evaluation of Visual Information Preservation}

To verify whether the compressed CoTs generated by XMCC retain critical visual information, we employ an LLM-as-Judge evaluation protocol. Specifically, we prompt a capable MLLM (i.e., Qwen3-VL-8B) to assess each compressed CoT based on two key aspects: (1) whether it references essential visual details or salient elements present in the image; (2) whether it includes descriptions of spatial relationships. The judge model then assigns an integer score between 1 and 5, reflecting the frequency and fidelity with which such visual content appears in the compressed reasoning.
As shown in Table~\ref{tab:vis-expl}, XMCC consistently outperforms all baseline methods in this evaluation. Notably, XMCC achieves a score that is 1.12 points higher than that of StepEntropy. This finding suggests that text-based, step-level compression methods may inadvertently discard crucial spatial or visual cues during compression when applied directly to multimodal reasoning. In contrast, XMCC explicitly accounts for visual content during the compression process, enabling it to preserve key cross-modal evidence and thereby enhance multimodal reasoning performance.

\subsection{Evaluation of Explanation Quality}

To assess the explanatory capability of XMCC, we evaluate the quality of the generated explanations. Specifically, we prompt a LLM (i.e., Qwen3-8B) to rate each explanation on a 1–5 integer scale based on three criteria: (1) whether it includes an inspection of the input CoT; (2) whether it articulates the rationale behind compression decisions; (3) whether it analyzes redundant or superfluous reasoning steps. The final score for each sample is the LLM's holistic rating, and we report the average score across the entire dataset. 
Besides, since existing compression methods generally lack explanations for the CoT compression, we use the raw LLM without any CoT compression policy training to generate compressed CoTs and their explanations (i.e., No Training) as our baseline. As shown in Table~\ref{tab:vis-expl}, XMCC achieves an average explanation quality score that is higher than this baseline by 1.72, providing strong evidence for the effectiveness of XMCC in generating meaningful explanations.

\vspace{-0.1cm}
\subsection{Ablation Study}
To assess the contribution of individual components in the XMCC framework, we conduct ablation studies. Specifically, during RL training, since the format reward and the outcome reward are indispensable, we ablate the step-wise criticality reward $R_{\text{step}}$ in Eq.~(\ref{overall-eq}). The results are reported in Table~\ref{tab:ablation}. We observe that removing $R_{\text{step}}$ leads to a performance degradation, indicating that $R_{\text{step}}$ plays a crucial role as a direct supervisory signal for the quality of the refined reasoning chain. This also highlights the necessity of monitoring CoT validity during compression to preserve downstream task performance.
Meanwhile, the average CoT length remains at a relatively low level after removing $R_{\text{step}}$, indicating that the remaining length reward $R_{\text{len}}$ alone is sufficient to effectively regulate the length of compressed CoTs and thereby ensure efficient inference.

\begin{table}[t]
\centering
\caption{Abation studies on MathVista and WeMath.}
   \renewcommand\arraystretch{.9}
\setlength{\tabcolsep}{4.5mm}{
    \scalebox{0.75}{
\begin{tabular}{lccccc}
\toprule
\multirow{2}{*}{Method} &  
\multicolumn{2}{c}{\textbf{MathVista}} &
\multicolumn{2}{c}{\textbf{WeMath}} \\
\cmidrule(lr){2-3} \cmidrule(lr){4-5}
 & Acc. & AvgLen & Acc. & AvgLen \\
\midrule
\multicolumn{3}{l}{\small{\textit{\textbf{SFT on Qwen2.5-VL}}}} \\
w/o $R_{step}$  & 34.8 &  72.4 & 51.6 &  83.1 \\
XMCC            & 37.6 &  89.7 & 54.9 &  99.6 \\
\midrule
\multicolumn{3}{l}{\small{\textit{\textbf{SFT on Qwen3-VL}}}} \\
w/o $R_{step}$  & 42.6 &  75.1 & 59.7 &  80.2 \\
XMCC            & 45.3 &  91.0 & 63.0 &  97.7 \\
\bottomrule
\end{tabular}
}}
\label{tab:ablation}
\vspace{-0.4cm}
\end{table}
\subsection{Analysis of Input CoT Quantity}
As discussed in section~\ref{sec:syntrajs}, using multiple input CoTs enables the compressor to exploit diverse reasoning perspectives, thereby improving the quality of the compressed CoTs. To validate this, we construct compressed datasets with $N=1$, $3$, and $6$ input CoTs per sample, and then SFT the model and evaluate downstream task performance. As shown in Figure~\ref{fig:combo}, both accuracy and reasoning efficiency improve as $N$ increases. When $N=6$, Qwen3-VL based XMCC   achieves over a $4\%$ accuracy gain compared to the case with $N=1$. These results indicate that multi-trajectory inputs effectively enhance compression quality and, consequently, significantly improve task performance.

\section{Conclusion}
In this paper, we proposed XMCC, an explainable multimodal CoT compression framework that enhanced reasoning efficiency by removing redundant steps while providing explanations of which information was preserved or discarded. Specifically, XMCC formulated compression as a sequential decision-making process optimized via reinforcement learning, guided by a multi-component reward function. By selectively retaining essential reasoning steps and cross-modal alignment cues, XMCC simultaneously compressed CoTs and provided natural-language explanations for each compression decision. Extensive experiments demonstrated that these compressed CoTs could be used to effectively fine-tune base models, resulting in a multimodal reasoner that achieved both high inference efficiency and reliable task performance.

\section*{Impact Statement}
This paper presents work whose goal is to advance the field of machine learning. There are many potential societal consequences of our work, none of which we feel must be specifically highlighted here.

\bibliography{example_paper}
\bibliographystyle{icml2026}

%%%%%%%%%%%%%%%%%%%%%%%%%%%%%%%%%%%%%%%%%%%%%%%%%%%%%%%%%%%%%%%%%%%%%%%%%%%%%%%
%%%%%%%%%%%%%%%%%%%%%%%%%%%%%%%%%%%%%%%%%%%%%%%%%%%%%%%%%%%%%%%%%%%%%%%%%%%%%%%
% APPENDIX
%%%%%%%%%%%%%%%%%%%%%%%%%%%%%%%%%%%%%%%%%%%%%%%%%%%%%%%%%%%%%%%%%%%%%%%%%%%%%%%
%%%%%%%%%%%%%%%%%%%%%%%%%%%%%%%%%%%%%%%%%%%%%%%%%%%%%%%%%%%%%%%%%%%%%%%%%%%%%%%
\newpage
\appendix
\onecolumn

\section{Details of Datasets}
To comprehensively evaluate the impact of CoT compression on multimodal reasoning, we construct XMCC-Dataset. The seed data for XMCC-Dataset is sampled from two sources: Geo170k and ScienceQA. Geo170k is a visual reasoning dataset encompassing mathematical reasoning, geometric analysis, and diagram-based problem solving, while ScienceQA is a multimodal scientific reasoning dataset that integrates images, questions, and domain-specific knowledge across physics, chemistry, and biology.

In constructing XMCC-Dataset, we first sample 9,000 image–question–answer (IQA) triples from Geo170k and ScienceQA. For each IQA triple, we generate diverse CoTs using three heterogeneous MLRMs: Qwen3-VL~\cite{qwen3-vl}, InternVL3.5~\cite{internvl3_5}, and Kimi-VL~\cite{kimi-vl}. For each model, we perform two independent inferences with different random seeds, resulting in 6 distinct CoTs per instance (i.e., $M=3$, $K=2$, $N = M \times K = 6$).

Finally, we split the synthesized data into two parts: $\mathcal{D}_{\text{train}}$ for compressor training, and $\mathcal{D}_{\text{com}}$, which serves as the dataset to be compressed. Applying the trained compressor to $\mathcal{D}_{\text{com}}$ yields the supervised fine-tuning data $\mathcal{D}_{\text{sft}}$. Statistics of XMCC-Dataset are summarized in Table~\ref{tab:dataset_stats}.

\begin{table}[h]
\centering
\caption{Statistics of the XMCC-Dataset, where $\mathcal{D}_{\text{train}}$ is the training data, $\mathcal{D}_{\text{com}}$ is the dataset to be compressed, and $\mathcal{D}_{\text{sft}}$ is the compressed dataset.
}
\begin{tabular}{lc}
\toprule
\textbf{Item} & \textbf{Value} \\
\midrule
\#Samples & 9,000 \\
$|\mathcal{D}_{\text{train}}|$ & 3,000 \\
$|\mathcal{D}_{\text{com}}|$ & 6,000 \\
$|\mathcal{D}_{\text{sft}}|$ & 6,000 \\
CoT length of $\mathcal{D}_{\text{sft}}$ ($\mu\pm\sigma$) & 104.6 $\pm$ 55.8 \\
\bottomrule
\end{tabular}
\label{tab:dataset_stats}
\end{table}

\section{Details of Benchmarks}

To comprehensively access the effectiveness of CoT compression, we evaluated the fine-tuned model across multiple multimodal reasoning benchmarks, including MathVista~\cite{lu2023mathvista}, WeMath~\cite{qiao2024we}, MMStar~\cite{chen2024mmstar}, MMMU~\cite{yue2024mmmu} and R1-Onevision-Bench~\cite{yang2025r1}:

\paragraph{Benchmarks}
\begin{itemize}
    \item MathVista~\cite{lu2023mathvista}: a mathematical benchmark constructed to integrate challenges across diverse mathematical and visual tasks. Its Test Mini split, containing approximately 1,000 samples, is utilized in our evaluation. 
    \item WeMath~\cite{qiao2024we}: a benchmark designed to explore problem-solving mechanisms beyond end-to-end performance. We adopt its Test Mini split, comprising around 1,740 samples, with average accuracy serving as our primary reporting metric. 
    \item MMStar~\cite{chen2024mmstar}: a challenging multimodal benchmark that evaluates fine-grained perception, spatial reasoning, and complex cross-modal alignment through carefully curated real-world images and questions. It contains 1,500 expert-annotated samples spanning diverse domains such as geometry, diagram interpretation, and scientific visualization.
    \item MMMU~\cite{yue2024mmmu}: a massive multimodal benchmark covering 30 subjects across six disciplines (e.g., STEM, humanities, social sciences), requiring college-level knowledge and sophisticated multimodal understanding. We use acombination of its dev and validation set and report average accuracy.
    \item R1-Onevision-Bench~\cite{yang2025r1}: a comprehensive multimodal reasoning benchmark covering mathematics, physics, chemistry, biology, and logical deduction across 38 subcategories. It comprises 942 problems paired with images. We adopt the full benchmark and report average accuracy as the primary metric. 
\end{itemize}

\section{Case Study on Fine-tuned Models}

\begin{figure}[b]
    \centering
    \includegraphics[width=1\linewidth]{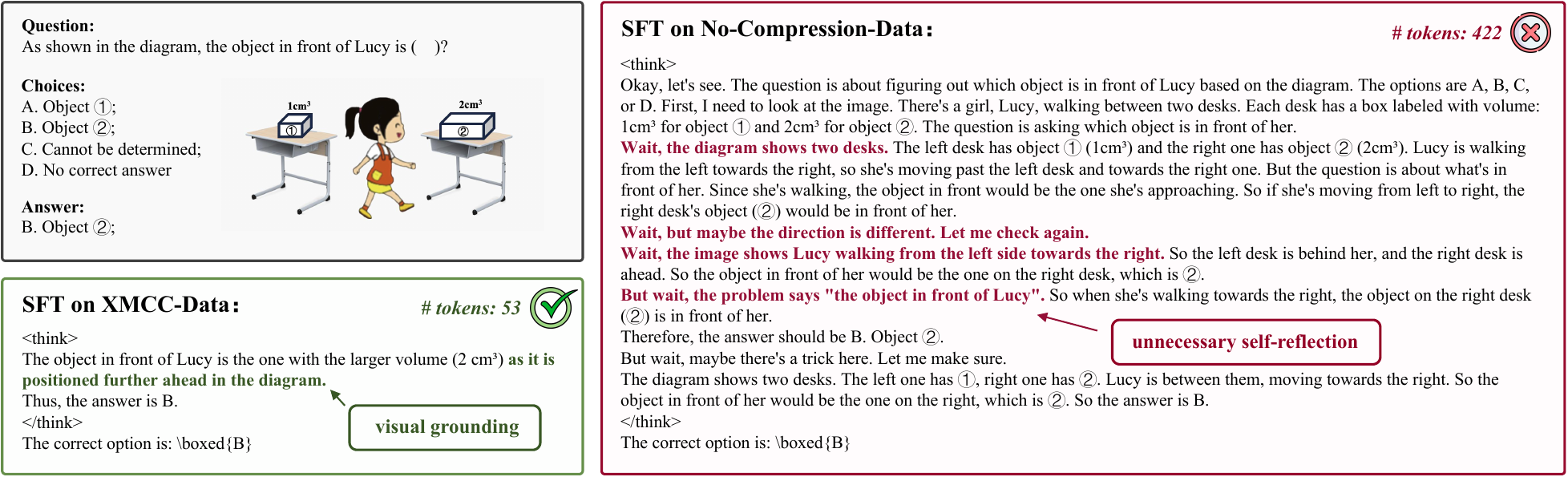}
    \caption{Case Study on SFT Models. Text in the box at the lower left corner is generated by the model fine-tuned on XMCC data, while text in right box is generated by the model fine-tuned on uncompressed CoTs.}
    \label{fig:sftcasestudy1}
\end{figure}

\begin{figure}
    \centering
    \includegraphics[width=1\linewidth]{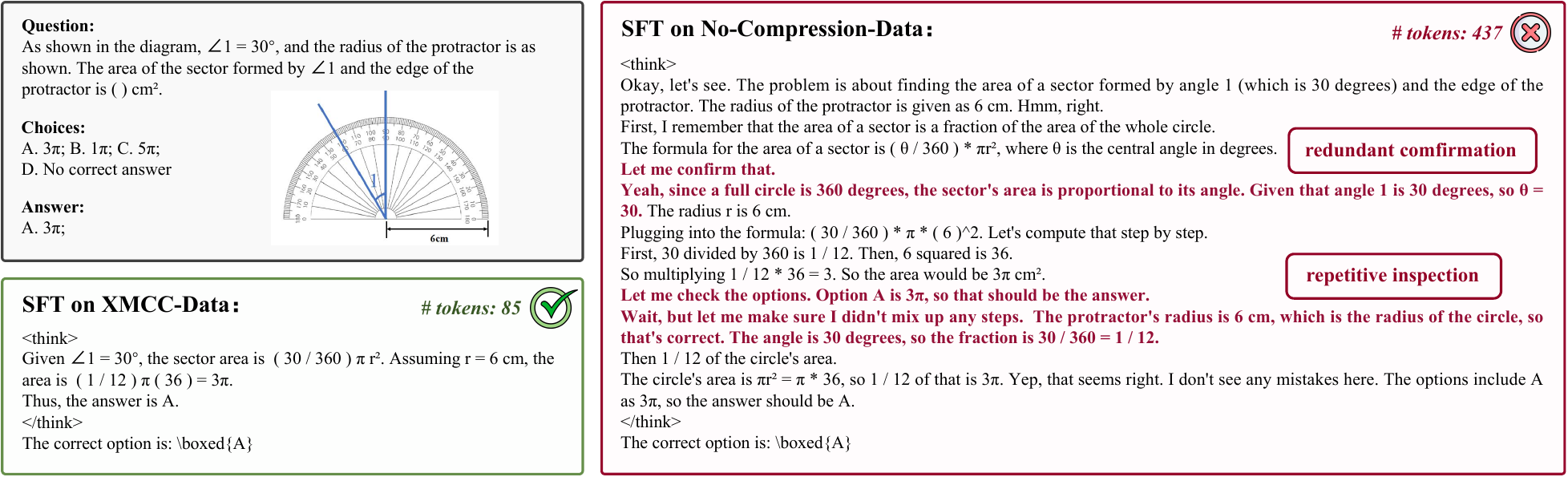}
    \caption{Case Study on SFT Models. Text in the box at the lower left corner is generated by the model fine-tuned on XMCC data, while text in right box is generated by the model fine-tuned on uncompressed CoTs.}
    \label{fig:sftcasestudy2}
\end{figure}

\begin{figure}
    \centering
    \includegraphics[width=1\linewidth]{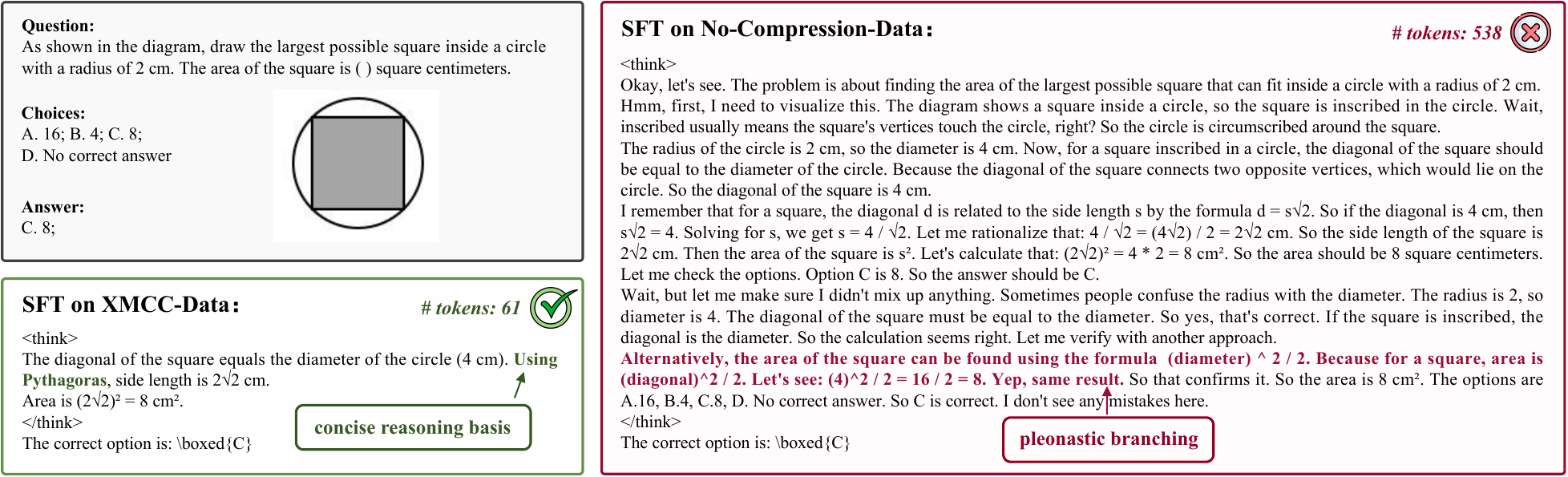}
    \caption{Case Study on SFT Models. Text in the box at the lower left corner is generated by the model fine-tuned on XMCC data, while text in right box is generated by the model fine-tuned on uncompressed CoTs.}
    \label{fig:sftcasestudy3}
\end{figure}

\begin{figure}
    \centering
    \includegraphics[width=1\linewidth]{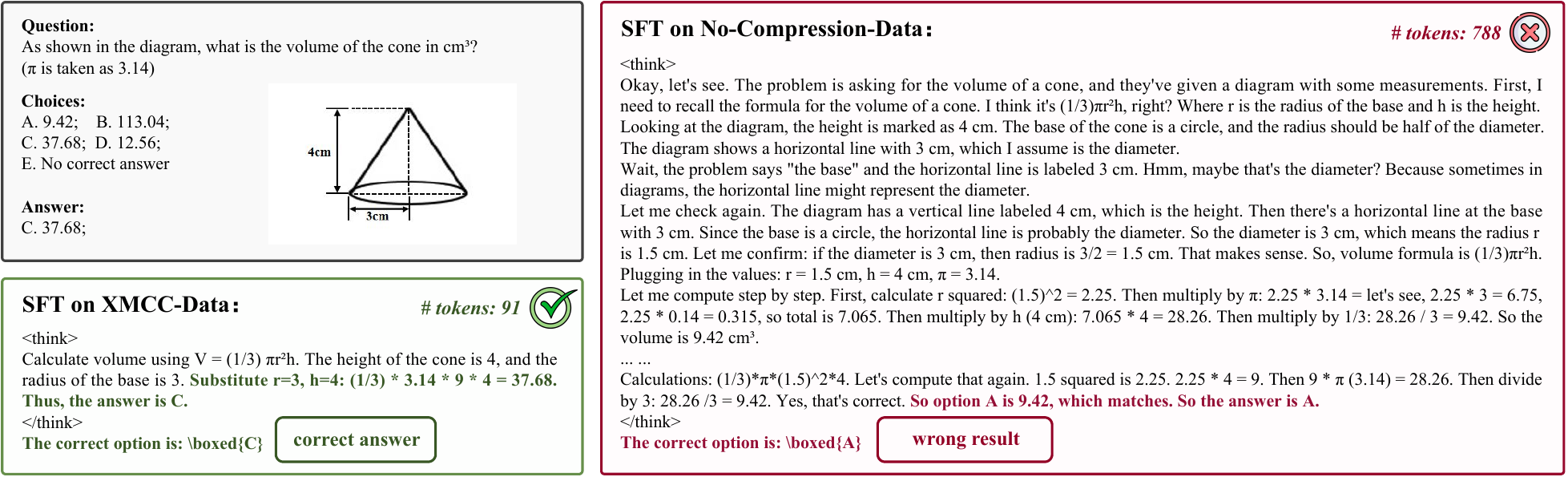}
    \caption{Case Study on SFT Models. Text in the box at the lower left corner is generated by the model fine-tuned on XMCC data, while text in right box is generated by the model fine-tuned on uncompressed CoTs.}
    \label{fig:sftcasestudy4}
\end{figure}

To further investigate the impact of XMCC-compressed data on downstream reasoning performance, we conducted a case study comparing models fine-tuned on different data. As shown in Figures~\ref{fig:sftcasestudy1},~\ref{fig:sftcasestudy2},~\ref{fig:sftcasestudy3},~\ref{fig:sftcasestudy4}, models fine-tuned with refined data produce significantly shorter reasoning chains compared to baseline methods. Moreover, redundant step re-verification and unnecessary self-reflection are notably reduced. This clearly demonstrates the effectiveness of XMCC’s compression, enabling the model to generate more concise and focused reasoning processes.

\section{Model Prompts}

\begin{tcolorbox}[
colframe=black,
width=17.3cm,
arc=2mm, auto outer arc,
title={Prompt of the Compressor}, breakable,
]
\#\#\# Task \\
You are an expert in reasoning and analysis. Below, you will be given a question, its ground-truth answer, and several existing chains of thought (CoTs) on this problem. \\
 \\
Your goal is to generate a new CoT that is: \\
- Correct: Logically sound and consistent with the ground truth, \\
- Effective: Clearly captures the essential reasoning needed to solve the problem. Preserves all critical inference steps required to reach the correct conclusion; no key insight should be lost. \\
- Concise: Remove redundancy, merge steps, and eliminate tangential \\
 \\
You may draw insights from the provided CoTs but should aim to synthesize a more efficient and insightful reasoning path. \\
 \\
\#\#\# Output Format \\
1. Thinking Process: Enclose your entire internal thinking process within ``$<$think$>$'' and ``$<$/think$>$''. Every Thinking Process must start with "Okay, let's check these provided CoTs: ", with no exception. Explain how you identify redundant parts, preserve essential logic, ensure alignment with the ground truth and so on. In this section, you must analyze the provided CoT and explicitly justify why you removed or retained specific content - do not merely analyze the problem itself. \\
2. Refined CoT: Immediately after, provide your new CoT inside ``$<$refinement$>$'' and ``$<$/refinement$>$''.   \\
   - Use exactly two newline characters (``\textbackslash n\textbackslash n'') to separate distinct semantic segments.   \\
   - Do not include any extra text outside the tags. \\
3. Final Answer: Conclude with ``$<$answer$>$\textbackslash boxed\{...\}$<$/answer$>$'', where ``...'' is the exact answer choice (e.g., A, B, C, D). \\
 \\
----- Example ----- \\
$<$think$>$ \\
Okay, let's check these provided CoTs: ... The original CoTs repeat the same calculation three times and include irrelevant examples. ... I will retain only the core algebraic step that links the given equation to the solution. Look at the provided CoT, some steps are redundanct... ... (your thinking process)... ... \\
$<$/think$>$ \\
$<$refinement$>$ \\
Okay, ... ... Given the equation $2x + 4 = 10$, subtract 4 from both sides: $2x = 6$. \\
Divide by 2: $x = 3$. \\
$<$/refinement$>$ \\
$<$answer$>$ \textbackslash boxed\{A\}$<$/answer$>$ \\
------------------- \\
 \\
\#\#\# Question \\
\{Question\} \\
 \\
\#\#\# Ground Truth \\
\{Ground Truth\} \\
 \\
\#\#\# CoTs \\
\{CoTs\} \\
 \\
\#\#\# Your Response \\
\end{tcolorbox}

\begin{tcolorbox}[
colframe=black,
width=17.3cm,
arc=2mm, auto outer arc,
title={Prompt of the Visual Information Preservation Evaluation}, breakable,
]
\#\#\# Task \\
You are an expert in evaluating visual grounding in reasoning chains. Given a chain of thought (CoT), assess how effectively it preserves critical visual information from the associated image. Focus on two aspects: (1) references to salient visual elements (e.g., object attributes, distinctive features); (2) descriptions of spatial relationships (e.g., position, layout, relative arrangement). \\
Crucially, base your evaluation on the density of visual evidence rather than its absolute count. A concise CoT that embeds even a limited number of precise visual anchors should receive a high score due to its high information density. Conversely, an excessively long CoT should be penalized if its visual descriptions remain sparse relative to its length, that is, when visual grounding does not scale proportionally with verbosity. Prioritize reasoning that efficiently integrates visual cues without unnecessary elaboration. \\
 \\
Scoring guidelines: \\
5: Dense, precise visual grounding; every segment meaningfully references visual evidence \\
4: Strong visual anchoring with appropriate density; key spatial relations preserved \\
3: Moderate visual grounding; some redundancy or minor omissions \\
2: Sparse visual references; critical spatial cues missing despite verbosity \\
1: Almost no visual grounding; purely textual reasoning disconnected from the image \\
 \\
\#\#\# Output Format \\
Provide a brief analysis, then output the final score strictly as: \\
 \\
Score: \textbackslash{}boxed\{X\} \\
 \\
where X is an integer between 1 and 5. \\
No additional characters, spaces, or text may follow the closing brace. This format must be strictly followed. Otherwise, subsequent parsing will fail.\\
 \\
\#\#\# Examples \\
\{Examples\} \\
 \\
\#\#\# CoT to Evaluate \\
\{CoT\} \\

\#\#\# Your Evaluation \\
\end{tcolorbox}

\begin{tcolorbox}[
colframe=black,
width=17.3cm,
arc=2mm, auto outer arc,
title={Prompt of the Explanation Quality Evaluation}, breakable,
]
\#\#\# Task \\
You are an expert in evaluating explanation quality for reasoning compression. \\
Given an explanation generated during CoT compression, assess its quality based on three criteria: \\
(1) whether it explicitly inspects and analyzes the content of the input CoT; \\
(2) whether it clearly articulates the rationale behind specific compression decisions; \\
(3) whether it identifies and justifies the removal of redundant or superfluous reasoning steps. \\
Explanations that merely analyze the original question without examining the input CoT should receive low scores. \\
High-quality explanations should demonstrate awareness of the original reasoning trajectory and provide concrete justifications for pruning choices. \\
 \\
Scoring guidelines: \\
5: Comprehensive analysis of input CoT with precise rationale for each compression decision \\
4: Clear inspection of input CoT and sound justification for major pruning choices \\
3: Basic analysis of input CoT with partial or vague compression rationale \\
2: Superficial analysis focusing primarily on the question rather than the CoT \\
1: No meaningful analysis of input CoT; explanation disconnected from compression decisions \\
 \\
\#\#\# Output Format
Provide a brief analysis, then output the final score strictly as: \\
 \\
Score: \textbackslash{}boxed\{X\} \\
 \\
where X is an integer between 1 and 5. \\
This format must be strictly adhered to. \\
Otherwise, the subsequent parsing will fail. \\
 \\
\#\#\# Examples \\
\{Examples\} \\
 \\
\#\#\# Explanation to Evaluate \\
\{Explanation\} \\
 \\
\#\#\# Your Evaluation
\end{tcolorbox}

% \newpage

%%%%%%%%%%%%%%%%%%%%%%%%%%%%%%%%%%%%%%%%%%%%%%%%%%%%%%%%%%%%%%%%%%%%%%%%%%%%%%%
%%%%%%%%%%%%%%%%%%%%%%%%%%%%%%%%%%%%%%%%%%%%%%%%%%%%%%%%%%%%%%%%%%%%%%%%%%%%%%%

\end{document}